\documentclass[sigconf]{acmart}

\usepackage{amssymb}
\usepackage{pifont}
\RequirePackage[utf8]{inputenc}
\RequirePackage[LGR,T1]{fontenc}
\RequirePackage[english,greek]{babel}
\RequirePackage{alphabeta,csquotes}

\AtBeginDocument{%
  }

\setcopyright{acmlicensed}
\copyrightyear{2025}
\acmYear{2025}
\usepackage{xcolor}
\definecolor{darkgreen}{rgb}{0.0, 0.5, 0.0}
\acmConference[Petra 2025]{June 23--25,
  2025}{Corfu, GR}

\usepackage{graphicx}  
\usepackage{multirow}

\begin{document}
\selectlanguage{english}

\title{A Transformer-Based Framework for Greek Sign Language Production using Extended Skeletal Motion Representations}

\author{Chrysa Pratikaki}
\email{chrisaprat@gmail.com}
\affiliation{
  \institution{School of ECE, National Technical University of Athens}
  \city{Athens}
  \country{Greece}
}

\author{Panagiotis Filntisis}
\email{pfilntisis@athenarc.gr}
\affiliation{
  \institution{Robotics Institute, Athena Research Center}
  \country{Greece}
}

\author{Athanasios Katsamanis}
\email{nkatsam@athenarc.gr}
\affiliation{
  \institution{Athena Research Center}
  \city{Athens}
  \country{Greece}
}

\author{Anastasios Roussos}
\email{troussos@ics.forth.gr}
\affiliation{
  \institution{Institute of Computer Science, Foundation for Research \& Technology - Hellas (FORTH)}
  \country{Greece}
}

\author{Petros Maragos}
\email{maragos@cs.ntua.gr}
\affiliation{
  \institution{School of ECE, National Technical University of Athens and Robotics Institute, Athena Research Center}
  \city{Athens}
  \country{Greece}
}


\begin{abstract}
Sign Languages are the primary form of communication for Deaf communities across the world. To break the communication barriers between the Deaf and Hard-of-Hearing and the hearing communities, it is imperative to build systems capable of translating the spoken language into sign language and vice versa. Building on insights from previous research, we propose a deep learning model for Sign Language Production (SLP), which to our knowledge is the first attempt on Greek SLP. We tackle this task by utilizing a transformer-based architecture that enables the translation from text input to human pose keypoints, and the opposite. We evaluate the effectiveness of the proposed pipeline on the Greek SL dataset Elementary23, through a series of comparative analyses and ablation studies. Our pipeline's components, which include data-driven gloss generation, training through video to text translation and a scheduling algorithm for teacher forcing - auto-regressive decoding seem to actively enhance the quality of produced SL videos.
\end{abstract}




\received{30 November 2024}

\maketitle

\section{Introduction}

To address communication barriers between the DHH (Deaf and Hard-of-Hearing) and the hearing communities, the field of Sign Language Processing has emerged at the intersection of linguistics, computer vision, and machine learning. Sign Language Processing encompasses a variety of tasks aimed at bridging the gap between DHH and hearing communities by enabling the automatic translation, and generation of sign language. The most critical components of an effective sign language system are Sign Language Translation (SLT), and Sign Language Production (SLP). In this paper, we primarily focus on Sign Language Production (SLP).

SLP systems have primarily relied on basic animation techniques or rule-based models, which often fail to capture the subtleties of human motion and natural language. Recent advancements in deep learning, particularly neural machine translation models and generative networks, have opened new possibilities for generating more photorealistic sign language content. Despite these advances, current solutions are still in their early stages, with significant room for improvement in the fluidity and accuracy of the produced sign language videos. In this work, we utilize transformer architectures aiming to address the existing SLP limitations. Our contributions can be summarized as follows:


\begin{figure*}[htbp]
\centerline{\includegraphics[scale=0.41]{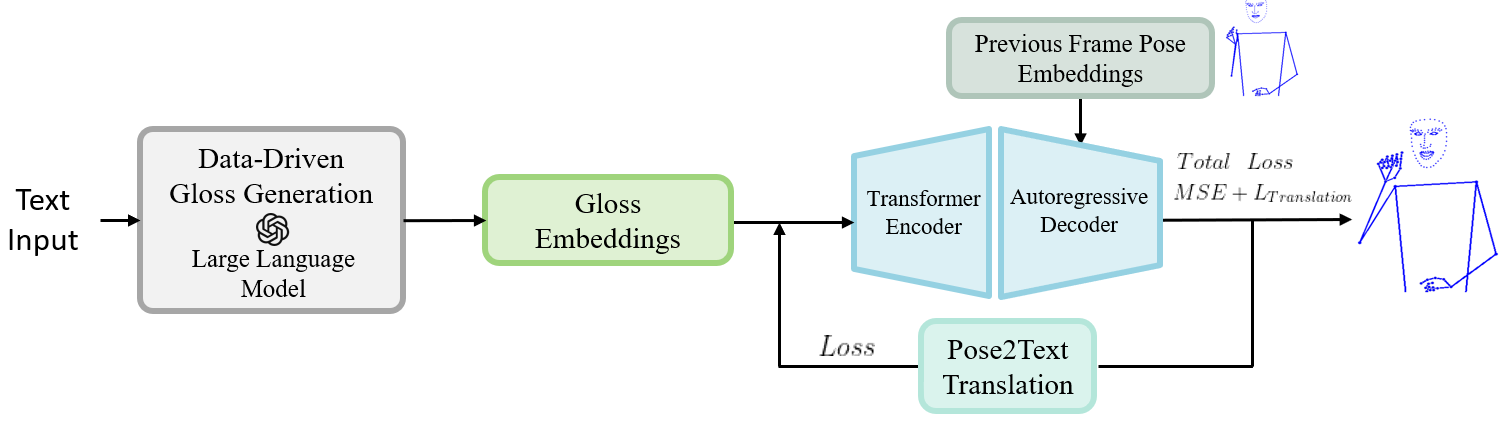} }
\caption{Overview of the proposed architecture: Given a \textbf{text} sentence as \textbf{input}, our SLP pipeline generates the corresponding sign language sequence. During training, the Encoder-Decoder structure learns through a sum of MSE Regression Loss (between frames) and CTC (pose-to-text) Loss. Optionally, training can happen using data-driven generated glosses to limit lexical diversity.}
\label{pipeline}
\end{figure*}

\begin{enumerate}
\item We present the first SLP system for the Greek Language based on deep learning.
    \item Our method incorporates components such as the use of a Pose-to-Text loss during training and SL gloss generation through text transcriptions, which help the quality of the generated sign poses. We also propose a scheduling algorithm that alternates between using teacher forcing and auto-regressive decoding during training.
    \item We conduct experiments on the publicly available Greek sign language dataset, 
 Elementary23 \cite{voskou2023new}. Through extensive quantitative evaluations we highlight the strengths and weaknesses of the proposed transformer-based architecture, achieving significant improvements on the quality of the SLP results.
\end{enumerate}

\section{Related Work} \label{sec:related}

\textbf{Sign Language Recognition and Translation:} Many recent works have tackled the task of Sign Language Recognition (SLR) and Sign Language Translation (SLT) using a variety of deep learning approaches, including RNNs (\cite{aslreccnnrnn}, \cite{Camgoz_2018_CVPR_rnn}), LSTMs (\cite{Camgoz_2017_ICCV}), GRUs (\cite{slt_keypoints}), and Transformers (\cite{Camgöz2020}, \cite{Camgoz2020multi}), after using CNNs for spatial feature extraction or Pose Estimation networks. Camgöz et al. \cite{Camgoz_2018_CVPR_rnn} formalized SLT as a sequence-to-sequence (seq2seq) learning problem. This approach employs CNNs for spatial feature extraction from sign language videos, which are then fed into an attention-based encoder-decoder framework to generate spoken language translations. These experiments were made on three different pipelines, gloss-tο-text, sign-to-text and sign-to-gloss-to-text, which uses gloss annotations as an intermediate layer. In another work, Camgöz et al. \cite{Camgöz2020} used transformer models for both the recognition  and translation pipelines. The encoders process sign video sequences to produce embeddings that capture both spatial and temporal features, while the decoders generate spoken language sentences. CTC Loss is used to facilitate learning without explicit alignment data, tying the recognition of sign glosses to the generation of text. Experimental results of the previously mentioned works prove that using gloss information as an intermediate step to spoken language translation improves the performance of the model, however relying on gloss annotations can be limiting  on larger datasets since they require professional annotation.

\textbf{Sign Language Production:} While Sign Language Translation (SLT) has seen considerable progress, Sign Language Production (SLP) remains under-explored, with a need for significant breakthroughs. Early works on SLP primarily relied on phrase lookup, direct sentence matching, and computer-generated avatar sign videos to produce realistic animated outputs, such as Tessa (BSL) \cite{tessa} and Simon (Sign Supported English) \cite{simon}. Recent advancements \cite{Saunders2020}, \cite{Stoll2020}, \cite{Saunders2022}, \cite{Stoll2022}, have redefined SLP as a \textbf{Neural Machine Translation} (NMT) task, leveraging sequence-to-sequence models to generate 2D pose sequences from text embeddings. Saunders et al. \cite{Saunders2020} pioneered a Transformer-based architecture for end-to-end SLP, employing a dual-transformer approach. Their Symbolic Transformer encodes text, while the Progressive Transformer generates continuous frame sequences, marking a significant step forward in automating and enhancing SLP. Several other works \cite{Saunders2022}, \cite{tze2023neural}, specialize on the photorealistic aspect of SLP and aim to synthesize realistic SL videos.


\textbf{Datasets:} Most mentioned works on SLT and SLP conduct their experiments on the publicly available \textbf{PHOENIX14T} dataset \cite{phoenix_koller15cslr}. This dataset includes a total of 8257 sequences performed by 9 signers along with their gloss annotations. Its relatively limited vocabulary of 1066 sign glosses allows for high quality SLT and SLP results. On the other hand the \textbf{Elemntary23} dataset \cite{voskou2023new}, is a recent GSL dataset, which contains annotations of the first three classes of Greek Elementary school books in all subjects. The Greek Language subset containts  9499 videos with a vocabulary of 14345 words, while the Maths subset containts 6583 videos with a vocabulary of 6457 words.


\textbf{Pose Estimation:}   Recent advances in the field of computer vision and pose estimation, make it possible to generate 2D or 3D landmarks from RGB images. Open Pose (\cite{openpose_cao2017realtime},\cite{openpose_wei2016cpm}) is one of the first and most popular frameworks for human pose estimation and is mostly used in the previously mentioned works on SLT and SLP. MediaPipe (MP) is an open-source framework for constructing multi-modal and cross-platform machine learning pipelines, supporting a broad range of applications, including pose and face detection. A particular implementation of MediaPipe is the MediaPipe Holistic model \cite{mediapipe_holistic}, which we particurally use in this work. MP Holistic employs a graph-based pipeline that processes different regions of interest (ROIs) within an image to estimate a total of up to 543 landmarks, which include 33 body pose landmarks, up to 468 facial landmarks, and 42 hand landmarks (21 per hand).

\section{Methodology}


Given a \textbf{text} sentence as \textbf{input}, our SLP pipeline generates the corresponding 2D sign language sequence in the form of extended skeleton representations, which include hand, face and body landmarks. An overview of the proposed architecture is presented in Fig. \ref{pipeline}. It consists of four main components: Feature Extraction, Gloss Extraction, Auto-regressive Decoding, and Pose-to-Text Translation, which are analyzed in the following sections.

\subsection{Feature Extraction}

\begin{figure}[htbp]
\centerline{\includegraphics[scale=0.085]{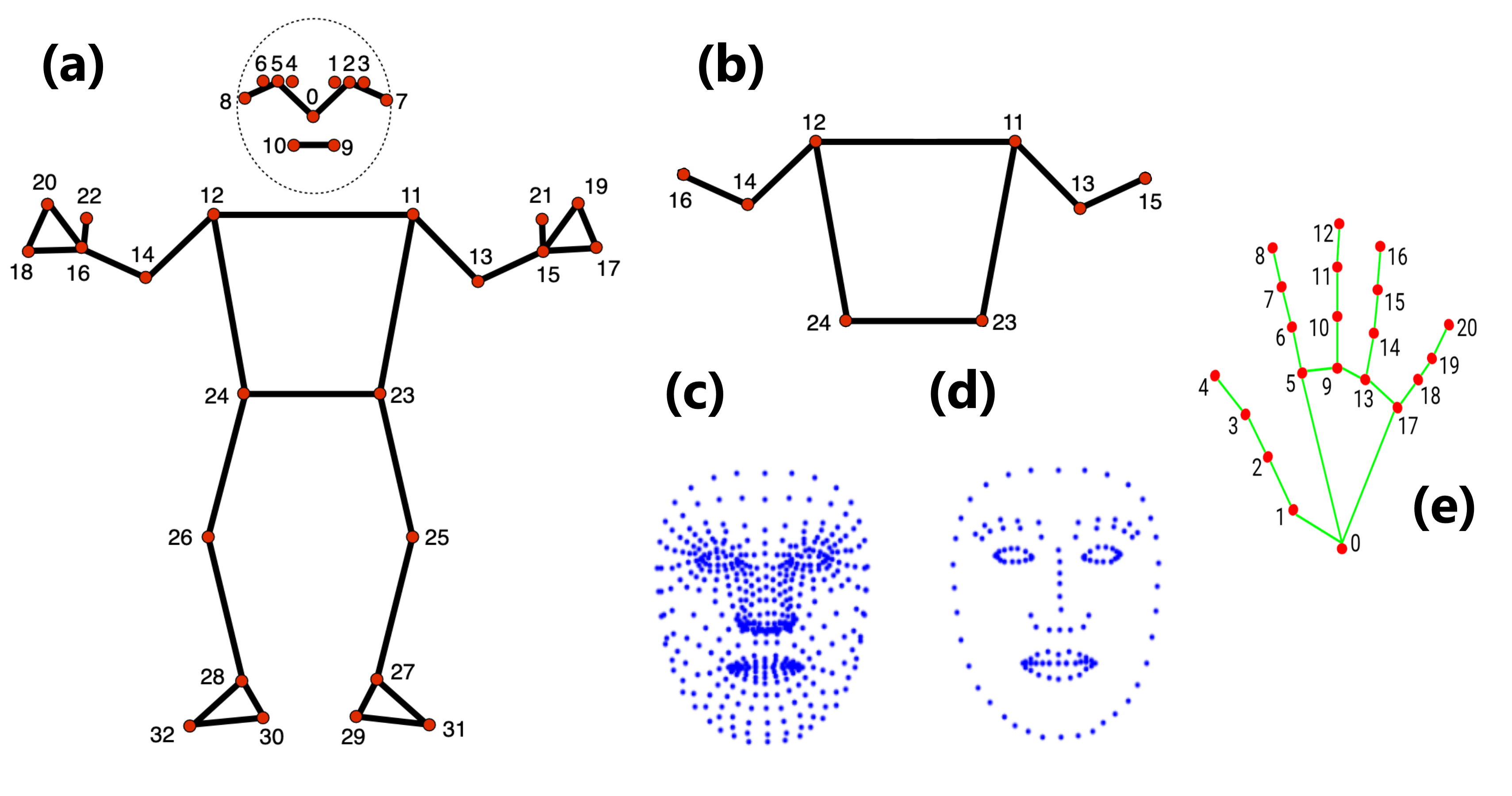}}
\caption{Extended Skeleton Representation based on MediaPipe Holistic \cite{mediapipe_holistic}: (a) Original 33 MP pose landmarks. (b) Selected 8 MP pose landmarks for SLP. (c) Original 478 MP face landmarks. (d) Selected 141 MP face landmarks for SLP. (e) MP hands.}
\label{mpposelands}
\end{figure}

We first extract the skeleton pose sequences from Elementary23 SL videos using MediaPipe (MP) Holistic \cite{mediapipe_holistic}. In order to expedite the training process, we sub-sample both the pose and face mesh landmarks. For the pose keypoints, we select the 8 points shown in Figure \ref{mpposelands} which include the body parts necessary for a SL video, such as the torso, elbows and wrists. For the face landmarks, we choose 141 instead of 468 face keypoints, which contain all the necessary face information, such as the mouth, eyes, nose and face perimeter. For each hand, we keep all 21 landmarks. This brings us to a total of 191 landmarks, instead of the original MP 543 landmarks, which is a substantial reduction to nearly one third.  Finally, the total extracted landmark sequence for each frame is defined as follows:
\begin{equation}
	\mathbf{P}_{f} = [\mathbf{a}_{left \ hand} || \mathbf{a}_{right \ hand} || \mathbf{a}_{face} || \mathbf{a}_{pose} || c_f]
\end{equation}

where $P_f$ is the landmark sequence for the f-th frame, $c_f$ is the counter value ranging from 0 to 1 that indicates the relevant frame posisition and $||$ the concatenation symbol.

\subsection{Text to Video Transformer Module}

\begin{figure}[htbp]
\centerline{\includegraphics[scale=0.2]{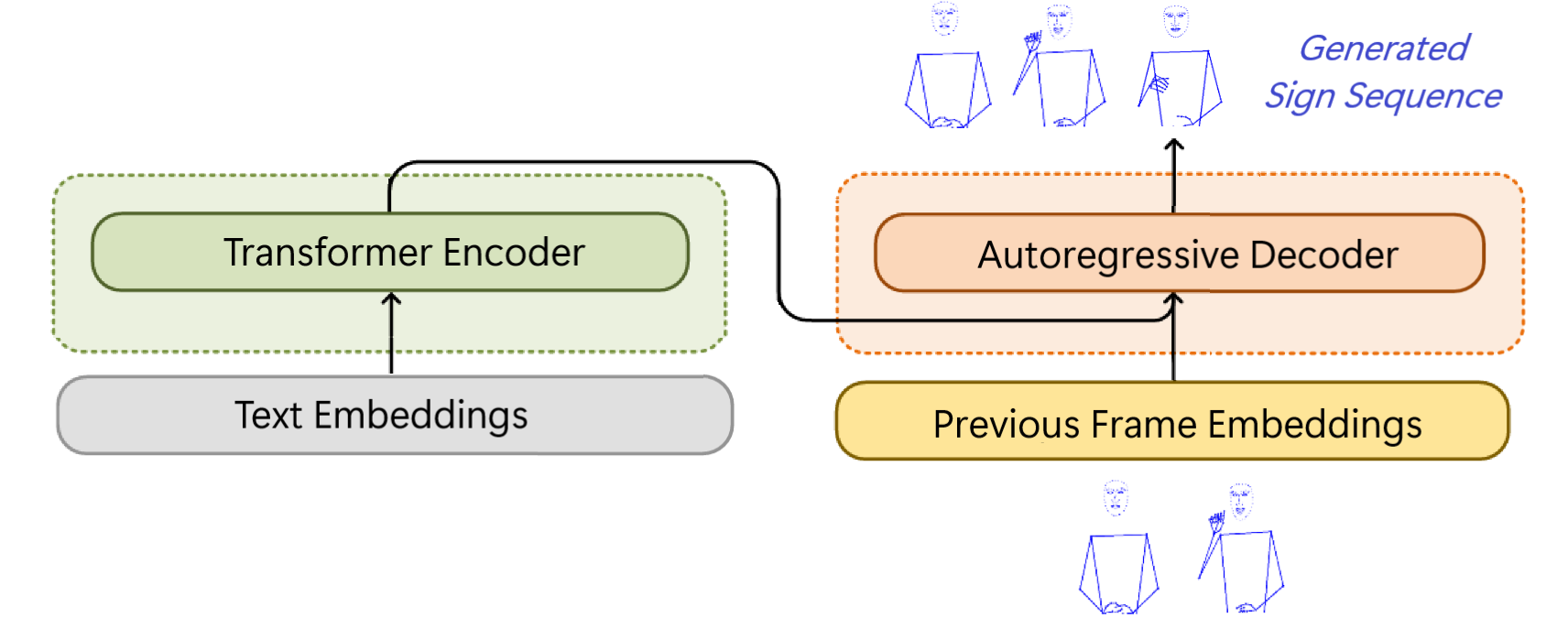} }
\caption{Proposed Sign Language Production Transformer}
\label{slp_back}
\end{figure}

After extracting the pose sequences from the dataset, we employ a transformer-based architecture to tackle the Sign Language Production (SLP) task, and ultimately transform text sentences to sign sequences. The key distinction between our approach and a classic encoder-decoder architecture is that the decoding process happens auto-regressively, meaning the model produces a sign pose frame at each time-step given the text embeddings and the previously generated pose embeddings. The training objective of the transformer module is concluded by regressively calculating the MSE between the ground-truth $y_{1:F}^{*}$ landmark sequence and the predicted landmark sequence $\hat{y}_{1:F}$, with $F$ being the total number of frames. Fig. \ref{slp_back} shows the \textbf{text2pose} Transformer architecture.

\subsection{Teacher Forcing vs Auto-regressive Decoding}

In previous methods \cite{Saunders2020}, transformer models were trained using teacher forcing. This approach involves providing the model with the ground truth spatial embeddings from the previous frame during sequence generation. By using the correct embeddings as input, this method enables parallel training with known outputs. While teacher forcing has demonstrated satisfactory results on limited vocabulary datasets such as PHOENIX14T \cite{Saunders2020, Camgöz2020}, it struggles with the broader and more diverse Greek Sign Language dataset. In general, while teacher forcing provides better training stability and ensures alignment between inputs and outputs—particularly in the earlier stages of training—it suffers from error compounding during inference, as the network is unable to recover from its own prediction errors.


On the contrary, autoregressive training generates frame sequences sequentially during training as well. In this approach, the model predicts each frame by conditioning on the spatial embeddings it has previously generated. Before applying the MSE loss, the entire sign sequence is generated from the text embeddings, effectively mimicking the inference process. This allows the model to learn to correct its own errors rather than relying on ground-truth inputs. However, this training process is considerably more time-consuming than teacher forcing due to the sequential nature of frame generation.

To balance efficiency and effectiveness, we employed a hybrid approach, training the model using teacher forcing and autoregressive generation for half of the epochs each. Specifically, we began training with teacher forcing to leverage its stability and strong input-output alignment during the critical early stages of training. This ensures the model effectively learns the foundational relationships in the data. We then switched to autoregressive training, allowing the model to learn to correct its own errors and better handle the challenges of inference. This strategy combines the strengths of both methods, resulting in improved performance compared to using either method independently. Quantitative results comparing these approaches are presented in Section \ref{sec:res}.




\subsection{Video to Text Translation Module}

An integral component of our training pipeline is the implementation of the \textbf{pose-to-text} loss. This approach entails the pre-training of a distinct Translation model that maps the 2d sign sequences to text, which is subsequently utilized during the training of the text2pose (forward process) model. The objective is to enhance accuracy and prevent the model from regressing to mean pose, which often happens when only training with MSE loss, and also prove its ability to reinforce the quality of the forward translation. The translation loss, essential for both training and evaluation, is formulated following \cite{Camgöz2020} as follows:
\begin{equation}
     L_T = 1 - \prod ^{U} _{u=1}  \sum ^{D} _{d=1} p(\hat{w}^d_u) p (w^d_u|h_u)
\end{equation}
where $p(\hat{w}^d_u)$ is the probability of word $w^d$ at decoding step u, while D is the vocabulary size. $ \prod ^{U} _{u=1}  p (w^d_u|h_u) $ is calculated by sequentially applying CTC Loss on a frame level for each
word.
\begin{figure}[htbp]
\centerline{\includegraphics[scale=0.15]{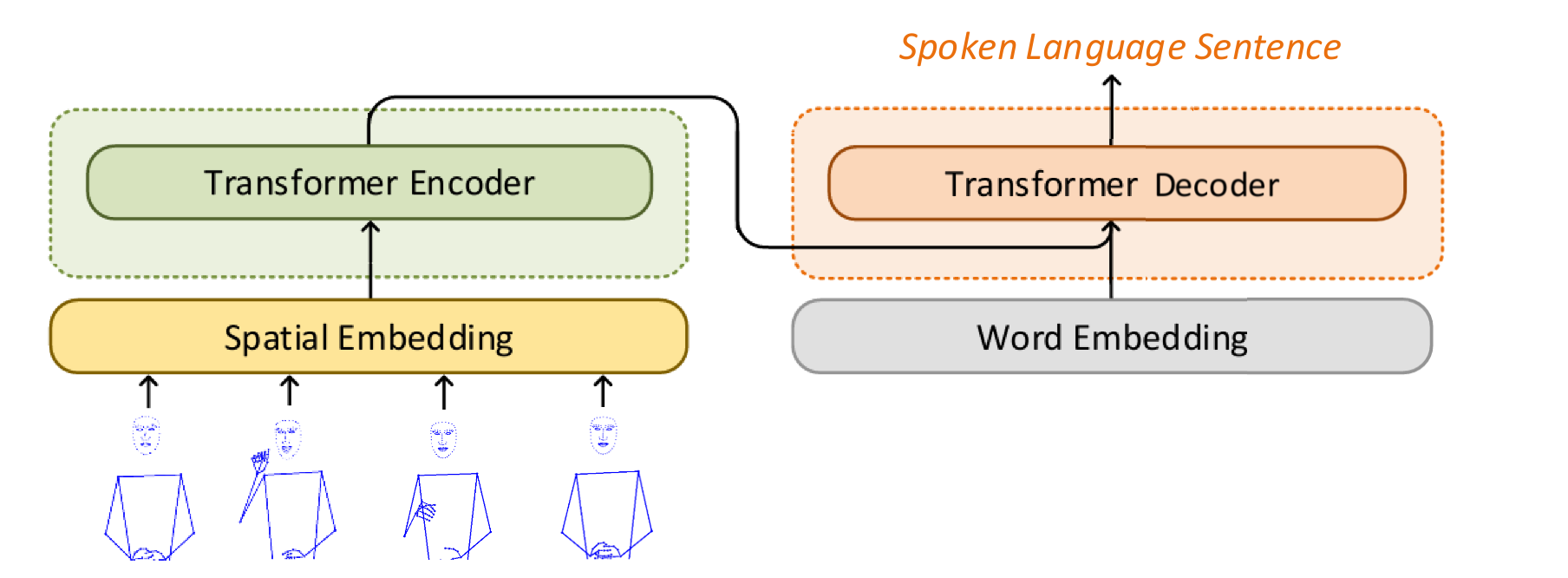} }
\caption{Proposed Sign Language Translation Transformer}
\label{slt_back}
\end{figure}

In order to implement the pose-to-text translation model we built on the state-of-the-art, publicly availiable network Sign Language Transformers by Camgoz et al. \cite{Camgöz2020}. Simplifying the overall training process that performs both SL recognition and translation, we keep solely the translation loss objective, aiming to achieve the desired results through a direct sign2text model. The complete architecture of the \textbf{pose-to-text} module is shown in Figure \ref{slt_back}.

\subsection{Gloss Extraction as an intermediate step}

Next, we explored the use of off-the-shelf large language models (LLMs) to automatically generate gloss annotations for the SL dataset. This approach effectively reduces the lexical diversity of the dataset by condensing commonly used words, such as articles and connective phrases, while preserving the overall meaning of the sentences.  Given the established benefits in previous literature, where gloss annotations as an intermediate step have been shown to enhance model performance \cite{Saunders2020}, \cite{Stoll2020}, we anticipate observing similar improvements in our experiments. Shown below in table \ref{glossexamples} as some Gloss Generation examples from the Elementary23 Dataset using the GPT-4o API.

\begin{table}[htbp]
\footnotesize
\begin{center}

\begin{tabular}{|p{1cm}|p{7cm}|}
\hline
Prompt & Transform this Greek sentence into Greek Sign Language gloss:  "I complete the table by first estimating the values approximately and then checking my calculations" \\
\hline
Gloss & COMPLETE TABLE ESTIMATE FIRST VALUES APPROXIMATELY CHECK THEN CALCULATIONS \\
\hline
Prompt & Transform this Greek sentence into Greek Sign Language gloss: "The line of symmetry divides shhapes into two equal parts" \\
\hline
Gloss &  LINE SYMMETRY DIVIDES SHAPE TWO EQUAL PARTS\\
\hline

\hline
Prompt & Transform this Greek sentence into Greek Sign Language gloss: "I observe and continue the patterns" \\
\hline

Gloss & OBSERVE CONTINUE PATTERNS\\
\hline
\end{tabular}
\caption{Examples of the text-to-gloss sequence translation that we adopt, based on LLMs. Note that the original sentences and gloss outputs are in Greek, however we present here the English translations.}
\label{glossexamples}
\end{center}
\end{table}


\section{Experiments and Evaluation} \label{sec:res}

In this section, we provide extensive comparative analyses and ablation studies on the Elementary23 dataset. First, we explore how the model performs on Signer-dependent subgroup of the Maths subset (for the two most frequently appearing Signers in Elementary 23 Maths subset). Next we explore how the model performs on entire sections of the dataset, with a class-related theme (i.e. Maths Subset, Greek Language Subset). Following \cite{Saunders2020}, \cite{Saunders2022}, we perform evaluation using the NLP metrics BLUE-4 and ROUGE-L, and also DTW (Dynamic Time Wrapping) for measuring similarity between the produced and ground-truth sign sequence.

\subsection{Dataset}

As mentioned in section \ref{sec:related}, the Elementary23 dataset \cite{voskou2023new} contains annotations of the first three classes of Greek Elementary school books in all subjects, with a large vocabulary exceeding 30,000 words. In our work, we focus on The Greek Language subset which contains 9499 videos with a vocabulary of 14345 words, and the Math subset which contains 6583 videos with a vocabulary of 6457 words. Specifically for the Math subset, we begin our evaluation in subsection \ref{sec:sign1}, by training the SLP pipeline on the two most prominent signers, referred to as Signer A and Signer B, who appear in 3,476 and 746 videos, respectively. Then, in subsection \ref{sec:sign2}, we generalize our training process across all signers to achieve a more holistic result. Table \ref{dataset} visualizes the size and vocabulary of each subset used.
\begin{table}[htbp]
\small
	\begin{center}
		\begin{tabular}{|c|c|c|c|} \hline  
			& &Videos &\# Words \\
			\hline
			
			 \multirow{2}{*}{Math} & \textbf{Signer A} & 3473 & 3654 \\ \cline{2-4}
			                       & \textbf{Signer B} & 746  & 1059 \\
			\hline
			 \multirow{2}{*}{Greek} & \textbf{Signer A} & 2467 & 4749 \\ \cline{2-4}
			                        & \textbf{Signer B} & 1927 & 3535 \\
			\hline
		\end{tabular}
		\caption{Elementary23 subsets}
        \label{dataset}
	\end{center}
\end{table}

\subsection{Evaluation of the sign-to-text module} \label{sec:text}

First, we evaluate the Sign Language Translation (sign-to-text) Module. The sign-to-text module is crucial for our pipeline, as it's used during SLP evaluation in the following sections, as well as during text-to-sign SLP training. Table \ref{entire} shows that we achieve a BLUE-4 score of 7.69 on the Math and 5.52 on the Greek subset, which is quite promising and close to the 6.67 mentioned in the original paper \cite{voskou2023new} on the SLT task.

\begin{table*}[htbp]
\footnotesize 
\begin{center}
\begin{tabular}{ccccc}
\hline
 & \multicolumn{2}{c}{\textbf{Dev}} & \multicolumn{2}{c}{\textbf{Test}} \\
 Sign-to-Text Method& \textbf{BLEU-4$\uparrow$}  & \textbf{ROUGE$\uparrow$}  & \textbf{BLEU-4$\uparrow$}  & \textbf{ROUGE$\uparrow$}   \\
\hline
Voskou et al. \cite{voskou2023new}, trained on entire Elementary23  &  6.67  & - &  5.69& -\\
Ours, trained on Elementary23 Math &  7.58 &   15.11 & 7.69&   15.26 \\
Ours, trained on Elementary23 Greek & 5.63 &   14.56 & 5.52&   14.23 \\
\hline
\end{tabular}

\caption{Evaluation of the \textbf{sign-to-text} module on the Elementary23 SL Dataset. Please not that results are not directly comparable due to differences in the test sets. }
\label{entire}
\end{center}
\end{table*}

\subsection{Evaluation of signer-specific training} \label{sec:sign1}

To begin our SLP evaluation, we performed experiments on the Mathematics subset of the Elementary23 dataset, on the two most frequently appearing signers, Signer A and Signer B. To assess the model’s ability to generalize across different signers, we performed two separate training sessions: one focused exclusively on Signer A and the other on Signer B. For evaluation, we alternated between their respective test sets to measure cross-signer performance. Results are shown in table \ref{specific}. We clearly see that the model fails to produce accurate signs when the "wrong" test set is used. These findings suggest that the model struggles to generalize across signers, likely due to differences in signing styles or vocabulary correlations unique to individual signers. To address this limitation, we proceed to train our models on larger sections of the Elementary23 dataset, emphasizing the need for more generalized training approaches.

\begin{table}[htbp]
\small
	\begin{center}
		\begin{tabular}{|c|c|c|c|c|}
			\hline
			 &  \multicolumn{2}{|c|}{\textbf{Test - Signer A}}&  \multicolumn{2}{|c|}{\textbf{Test - Signer B}}\\ 
			& BLEU-1$\uparrow$ &\textbf{BLEU-4$\uparrow$}    &  BLEU-1$\uparrow$ &\textbf{BLEU-4$\uparrow$}    \\
			\hline
			
			 \textbf{Train - Signer A}  &  17.05 &\textbf{5.02} &  5.93    &\textcolor{red}{0.00} \\
			 \textbf{Train - Signer B} &   6.29 &\textcolor{red}{1.18} &  21.87    &\textbf{6.69}  \\
	
			\hline 
		\end{tabular}
		
		\caption{Ablation Study on the Elementary23 Greek Language SL Dataset. Best-performing results are highlighted in bold, while failure scores in the case of swapped signer test scores are shown in red.}
        \label{specific}
	\end{center}
\end{table}

\subsection{Signer Independent Studies} \label{sec:sign2}

Next, we focus on conducting experiments on entire sections of the Elementary23 dataset, disregarding the fact that videos are filmed with different signers. We specifically choose the entire Math subset and the Greek Language subset. Our first ablation study, shown in table \ref{auto}, compares training with Teacher Forcing (TF), Auto-regressive Decoding (AD), and their combination (TF+AD), underlining the benefits of employing a hybrid approach. While Auto-regressive Decoding achieves significantly higher BLEU-4 and ROUGE scores (5.4 and 14.5 on the dev set, respectively) compared to Teacher Forcing (1.69 and 8.52), the hybrid TF+AD model provides balance between computational efficiency and predictive accuracy. Notably, the hybrid model achieves the highest overall performance, both in the Greek and Math subsets, validating the importance of alternating decoding strategies during training.
\begin{table}[htbp]
\footnotesize 
\begin{center}
\begin{tabular}{cccccc}
\hline
& & & Time/ & \multicolumn{1}{c}{\textbf{Dev}} & \multicolumn{1}{c}{\textbf{Test}} \\
 & Method & Epochs & Epoch (s) & \textbf{BLEU-4$\uparrow$} & \textbf{BLEU-4$\uparrow$}    \\
\hline

\multirow{3}{*}{\parbox[c]{2mm}{\rotatebox{90}{Greek}}} 
& Teacher Forcing, \cite{Saunders2020} & 2500  & 5 & 0.49  & 0.35 \\
& Autoregressive Dec & 2500 & 30 & 4.3 &  4.13\\
& TF + AD & 1250, 1250 & 5, 30 & {\textbf{4.67}} &  {\textbf{4.46}} \\
\hline

\multirow{3}{*}{\parbox[c]{2mm}{\rotatebox{90}{Math}}} 
& Teacher Forcing, \cite{Saunders2020} & 2500  & 5 & 1.69  & 1.46 \\
& Autoregressive Dec & 2500 & 30 & 5.4 &  5.3 \\
& TF + AD & 1250, 1250 & 5, 30 & {\textbf{5.69}} &  {\textbf{5.59}} \\
\hline
\end{tabular}

\caption{Ablation Study on the Elementary23 Greek (Top) and Math (Bottom) SL Dataset. Best TF+AD results are highlighted in bold. Teacher Forcing (TF) method can be considered equivalent to the Progressive Transformers (PT) work \cite{Saunders2020}.}
\label{auto}
\end{center}
\end{table}





Our next ablation study, shown in table \ref{glossa}, on the Elementary23 Greek Language Subset, shows that the inclusion of pose-to-text Loss and Gloss annotations possessively affects on performance. Although BLEU-4 scores improve independently (4.42 dev, 4.55 test), the combination with Gloss yields mixed results, slightly reducing BLEU-4 on the dev set (4.06) but improving on the test set (4.32). This interplay suggests that while gloss annotations simplify linguistic diversity, over-reliance on glosses can limit adaptability.

\begin{table}[htbp]
	\begin{center}
		\begin{tabular}{cccc}
			\hline
			& & \multicolumn{1}{c}{\textbf{Dev}} & \multicolumn{1}{c}{\textbf{Test}} \\
			$pose-to-text \ Loss$ & $Gloss$ & \textbf{BLEU-4$\uparrow$}    & \textbf{BLEU-4$\uparrow$}    \\
			\hline
			
			\ding{55}& \ding{55}  & \textbf{ 4.17 }&    4.15 \\
			\ding{55}& \ding{51} & 3.56 &    3.44 \\
			\ding{51}& \ding{55} & \textbf{4.42} &    \textbf{4.55} \\
			\ding{51}& \ding{51} & 4.06 &   \textbf{4.32}\\
			
			\hline
		\end{tabular}
		
		\caption{Ablation Study on the Elementary23 Greek Language Dataset. We highlight best performing scores in both dev and test sets.}
        \label{glossa}
	\end{center}
\end{table}

\begin{figure}[htbp]
\centerline{\includegraphics[scale=0.5]{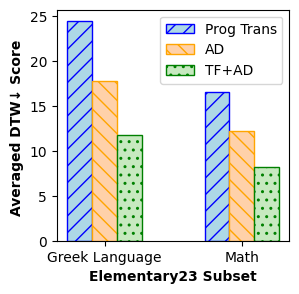} }
\caption{Comparison of the averaged DTW results on the Math and Greek Test Subsets. Again the hybrid combination of teacher forcing and auto-regressive decoding during training significantly improves sequence alignment.}
\label{RES}
\end{figure}

\begin{figure*}[htbp]
\centerline{\includegraphics[scale=0.354]{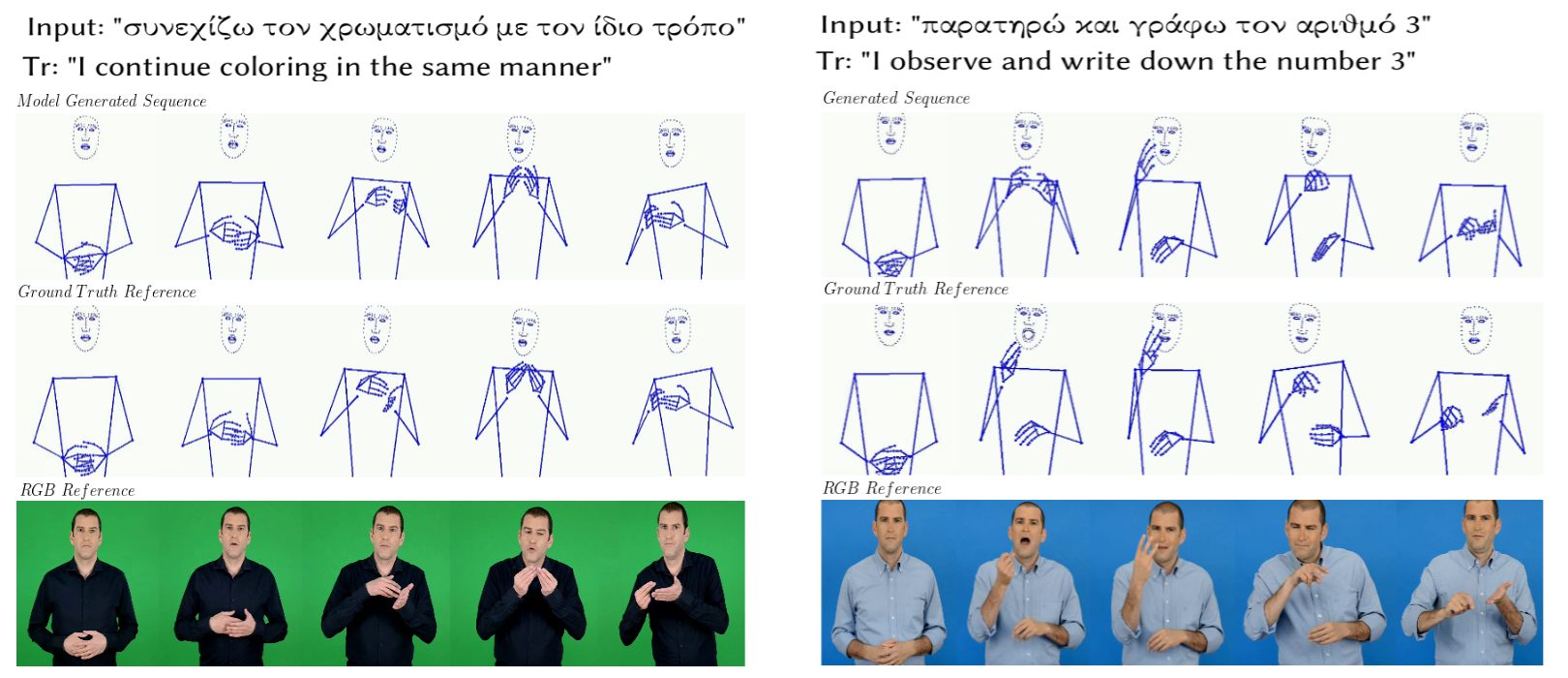} }
\caption{Sample (test set) visualizations of our SLP method. Top to bottom: Text inputs, 2D generated sign sequence from text embeddings, ground-truth sequence reference, RGB reference. Figures are best viewed in video form. }
\label{RES}
\end{figure*}

\begin{figure}[htbp]
\centerline{\includegraphics[scale=0.58]{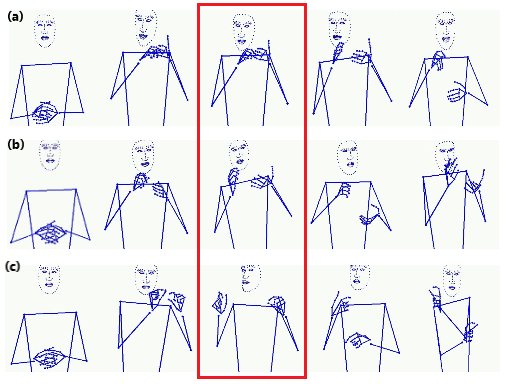} }
\caption{Sample visualization of the effect of the pose-to-text Loss. Top to bottom: \textbf{(a)} 2D Pose w/o pose-to-text Loss, \textbf{(b)} 2D Pose with pose-to-text Loss, \textbf{(c)} ground-truth sequence reference. When used, the generated poses show greater movement variability and regress less on mean pose.}
\label{RES}
\end{figure}

\subsection{Experimental Setup}

All models have been trained using 2-layer transformers with 4 attention heads,  embedding dimension 512. All SLP and SLT models used in a specific pipeline and data subset are trained on the same specifications for model compatibility. 

\subsection{Qualitative Results}

Lastly, in Figure \ref{RES} we showcase \textbf{representative results} from our proposed pipeline, evaluated on test sentences from the Elementary23 dataset using our best-performing models.

\section{Conclusions}

In this paper we presented the first SLP pipeline applied on Greek Sign Language Datasets, actively improving existing architectures through novel components. We presented our best results, which where achieved through the combination of gloss generation, decoding scheduling and pose-to-text translation training. These SLP methods find useful application mainly in the sign language learning process and education. In the future, we aim to expand our work so that it also incorporates a generative module for photorealistic SL video synthesis, as this is considered a necessary component for a SL user.  It's finally important to emphasize that SLP models are not intended to replace sign language interpreters. Instead, they serve as a complementary tool, providing an ethical and practical solution for educational purposes.

\bibliographystyle{ACM-Reference-Format}
\bibliography{sample-base}










\end{document}